%% file: FA_GANs_Facial_Attractiveness_Enhancement.tex
\documentclass[journal]{IEEEtran}
\ifCLASSINFOpdf
  \usepackage[pdftex]{graphicx}
  \graphicspath{{../pdf/}{../jpeg/}}
\else
\fi
\usepackage{amsmath}
\usepackage{amssymb}
\usepackage{algorithmic}

\usepackage{array}
\begin{document}
%
\title{FA-GANs: Facial Attractiveness Enhancement with Generative Adversarial
Networks on Frontal Faces}


\author{\IEEEauthorblockN{Jingwu He,
Chuan Wang,
Yang Zhang,
Jie Guo, and
Yanwen Guo}
\thanks{
J. He, Y. Zhang, J. Guo and Y. Guo are with the National Key
Laboratory for Novel Software Technology, Nanjing University, Nanjing
210023, China (e-mail: hejw005@gmail.com; yzhangcst@smail.nju.edu.cn; guojie@nju.edu.cn; ywguo@nju.edu.cn).

C., Wang is with the Megvii Technology (e-mail: wangchuan@megvii.com).}
}

%



\maketitle
\input{files/Abstract.tex}

\begin{IEEEkeywords}
Facial attractiveness enhancement, generative adversarial networks, geometry and appearance adjustment.
\end{IEEEkeywords}



%
\IEEEpeerreviewmaketitle

\input{files/Introduction.tex}

\input{files/RelatedWork.tex}
\input{files/Method.tex}

\input{files/Experiment.tex}
\input{files/Conclusions.tex}




%
%
%

\bibliographystyle{IEEEtran}
\bibliography{mm19}

%

%






\end{document}

%% file: files/Abstract.tex
\begin{abstract}
Facial attractiveness enhancement has been an interesting application of Computer Graphics and Vision over these years.
It aims to generate a more attractive face via manipulations of the facial image while preserving face identity.
In this paper, we propose the first Generative Adversarial Networks (GANs) for enhancing facial attractiveness in both the appearance and geometry aspects, which we call ``FA-GANs".
FA-GANs contain two parallel branches, which are both GANs used to enhance facial attractiveness in two perspectives: facial appearance and facial geometry.
With an embedded ranking module, the proposed FA-GANs are able to evaluate facial attractiveness and extract attractiveness features which are then further imposed on the enhancement results. This also enables us to train the networks without using paired faces as most previous methods have done.
Benefited from our parallel structure, the appearance and geometry networks are able to adjust face appearance as well as face geometry independently. The consistency of outcomes of the two branches are enforced by a geometry consistency module which links the two branches naturally.
To the best of our knowledge, we are the first to enhance facial attractiveness by considering both the appearance and geometry aspects using GANs under a unified framework.
Experimental results show that our FA-GANs generate compelling results and outperform the state-of-the-arts.
\end{abstract}

%% file: files/Introduction.tex
\section{Introduction}
\label{Introduction}
Related researches have shown that human faces tend to make a powerful first impression and thus may have potential influence on later social behaviors. The advantages brought by attractive faces are proven in many scientific studies~\cite{perrett1998effects}. Consequently, growing numbers of celebrities enhance their facial attractiveness in daily life.
\input{files/figure_teaser.tex}
Enhancing facial attractiveness is the process of adjusting a given face in the sense of visual aesthetics. Besides its academic purpose, it also shares a wide market in entertainment industry as a way to beautify the portrait. Facial attractiveness enhancement thus has received considerable attention in the Graphics and Vision communities.

During the last decade, the proposed facial manipulation methods can be roughly divided into two categories. The first category resorts to traditional Computer Graphics techniques which warp the input face to alter facial appearance. To enhance 2D facial image attractiveness, Leyvand et al.~\cite{Leyvand:2008:DEF:1399504.1360637} introduce a method by learning the distances among facial key points and adjusting the facial image with multilevel free-form deformation. Li et al.~\cite{Li_2015_CVPR} simulate makeup through adaptations of physically-based reflectance models.
For the 3D face models with both geometry and texture information, Liao et al.~\cite{6143936} propose a sequence of manipulations concerning symmetry, frontal face proportion, and facial angular profile proportion.
The methods in the other category generate the desired face via building the generative models by using deep learning.
The representative methods are the deep generative networks which have exhibited a remarkable capability in facial image generation.
BeautyGAN~\cite{Li:2018:BIF:3240508.3240618} adjusts facial appearance by transferring the makeup style from a given reference face to another non-makeup one. Though promising results are generated, this method mainly focuses on style transfer and cannot be generalized to automatic facial attractiveness enhancement easily. It is widely recognized that the geometry structure of the face plays a critical role in facial attractiveness.
However, to the best of our knowledge, till now there does not exist a unified framework for improving the attractiveness involving both the appearance and geometry aspects.

The purpose of this paper is to automate facial attractiveness enhancement in a unified deep learning framework which could not only alter the facial appearance but also improve the geometry structure.  A key challenge in our task is the lack of paired data which can be used for training. It is obviously very expensive to collect a large quantity of paired faces such that each pair is of the same individual before and after enhancement for supervised learning.
Another challenge is that people may define facial attractiveness in different ways. Their answers vary a lot when asked to explain this concept. However, there does exist a general criterion in some ways, such that most actresses or actors are recognized by the majority for their beauty or handsomeness. Therefore, it is possible to explore the relationship between the ordinary and attractive faces to make the enhancement task tractable, even though it might be difficult to build a set of explicit, explainable, and well acknowledged rules to define facial attractiveness.

In this paper, we propose the facial attractiveness enhancement Generative Adversarial Networks, called FA-GANs, to achieve attractiveness enhancement by altering the input face in both the appearance and geometry aspects.
Given the difficulty in collecting paired data, FA-GANs are trained on a dataset we collected which consists of  the attractive faces and ordinay ones. The faces in the two subsets are, however, uncorrelated. To learn with unpaired faces, we specifically design a novel attractiveness ranking module which could extract attractiveness features. Facial attractiveness can thus be evaluated with implicit rules in a data-driven manner. 


FA-GANs consist of two parallel branches used for the enhancement of facial geometry and appearance, respectively. Both of them are with the structure of GANs. For the geometry branch, we employ the facial landmarks to depict the geometry structure of a face.
For the appearance branch, the ranking module with the structure of VGG-16 is pre-trained, and the feature maps extracted are utilized to depict the attractiveness rules.
The ranking module determines the attractiveness domain that the ordinary faces should be enhanced.
Then, the networks of deep face descriptor~\cite{BMVC2015_41} are utilized by minimizing the distance of the facial images before and after enhanced.
This guarantees the enhanced face and the original input can be recognized as the same individual.
A newly designed geometry enhancement consistency module is proposed to combine these two branches, ensuring FA-GANs enhance the facial attractiveness in two aspects consistently.

To the best of our knowledge, we are the first to enhance the facial attractiveness with GANs by considering not only the appearance aspect but also the facial geometry. Our comprehensive experiments show that FA-GANs generate compelling perceptual results and outperform the state-of-the-arts.

In addition to the general framework, our paper also makes the following technical contributions.
\begin{itemize}
\item We propose FA-GANs for facial attractiveness enhancement with a newly designed geometry enhancement consistency module to automatically enhance the input face in both geometry and appearance aspects.
\item The pre-trained attractiveness ranking module is embedded in FA-GANs to learn the features of attractive faces via unsupervised adversarial learning, which does not require the training data to be organized as paired faces.
\item FA-GANs consist of two branches each of which is able to work independently to adjust either facial appearance or geometry.
\end{itemize}

%% file: files/figure_teaser.tex
\begin{figure}[tb]
\centering
\includegraphics[width=\linewidth]{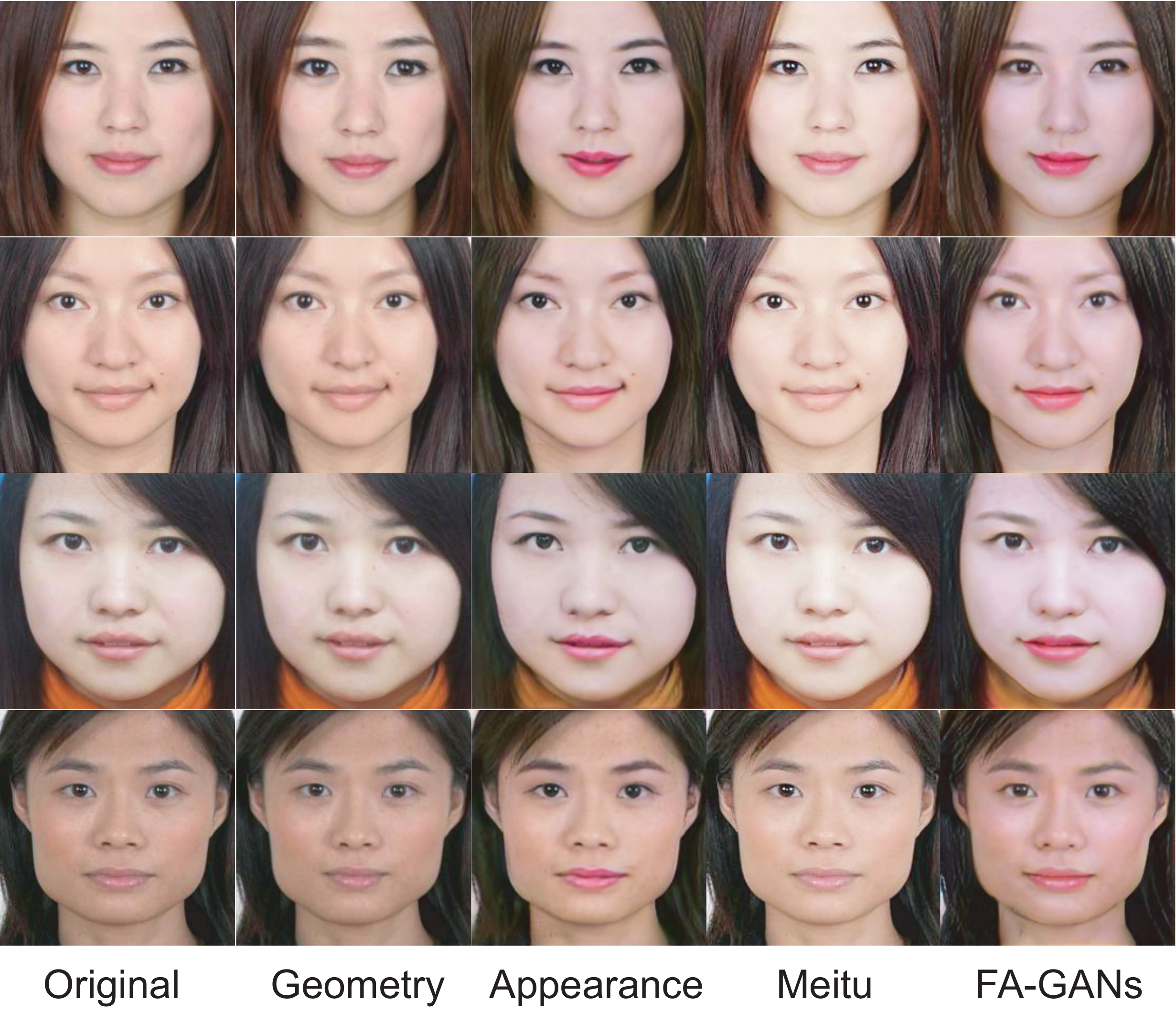}
\caption{Exemplar results of our FA-GANs and the mobile APP of Meitu for facial attractiveness enhancement. The original faces, geometry adjusted faces, appearance adjusted faces, Meitu adjusted faces, and our results are shown from left to right, separately.}	
\label{fig:teaser}	
\end{figure} 

%% file: files/RelatedWork.tex
\section{Related Work}
\label{Related Works}
Facial attractiveness has been investigated for a long time in many research areas, such as computer vision and cognitive psychology. Numerous empirical experiments suggest that beauty cognition does exist over the education background, age, and gender. Facial attractiveness enhancement is closely related to facial attractiveness analysis and face generation. In this section, we first review the researches in altering facial appearance and geometry, and then pay special emphasis on the recent facial image generation methods with generative adversarial networks.

\subsection{Facial Appearance Enhancement}
Arakawa et al. propose a system using interactive evolutionary computing which removes undesirable skin components from human facial images to make the face look beautiful~\cite{1595333}. Other works devote to making the enhancement via face makeup methods. Liu et al. propose a fully automatic makeover recommendation and synthesis system named beauty e-Expert~\cite{Liu:2013:WYS:2502081.2502126}. Deep learning has been used and shown remarkable performance. Liu et al. propose an end-to-end deep localized makeup transfer network to automatically synthesize the makeup for female faces, achieving natural-looking results~\cite{DBLP:conf/ijcai/LiuOQWC16}. Li et al. achieve makeup transfer from a given reference makeup face to the non-makeup one in high quality with their proposed BeautyGAN~\cite{Li:2018:BIF:3240508.3240618}. Chang et al.~\cite{Chang_2018_CVPR} propose PairedCycleGAN which can quickly transfer the style from an arbitrary reference makeup photo to an arbitrary source photo. Most of these methods based on deep learning focus on style transfer. By contrast, our aim is to automatically enhance facial image in the aspects of not only facial appearance but also face geometry. Given a target face, we do not require an additional image as the reference.

\subsection{Facial Geometry Enhancement}
The shapes of beautiful faces are defined in different ways by different individual groups, but beautiful faces can always be recognized to be attractive by individuals from other groups ~\cite{LAURENTINI2014184}.

For the 2D facial image, Leyvand et al. propose a data-driven approach for facial attractiveness enhancement~\cite{Leyvand:2008:DEF:1399504.1360637}, which adjusts the geometry structure by learning from the distance vectors of the landmark points. Li et al. propose a deep face beautification framework which is able to automatically modify the geometry structure of a face to boost the attractiveness~\cite{Li:2015:DFB:2733373.2807966}.
For the 3D face model, Liao et al. propose an enhancing approach by adjusting geometry symmetry, frontal face proportion, and facial angular profile proportion~\cite{6143936}. They apply proportion optimization to the frontal face restricted to the Neoclassical Canons and golden ratios. The combination of asymmetry correction and adjustment of frontal or profile proportions can achieve better results. However, the research also suggests that asymmetry correction is more effective than adjusting the proportions of the frontal or profile \cite{LAURENTINI2014184}. Qian et al. propose additive focal variational auto-encoder (AF-VAE)  to arbitrarily manipulate high-resolution face images~\cite{Qian_2019_ICCV}. Different from these methods which mainly alter face geometry based on empirical rules, we explore facial attractiveness by learning from unpaird training data in a data-driven manner, considering that unanimous and measurable rules about attractiveness are difficult to define explicitly.

\subsection{Generative Adversarial Networks}
Our facial enhancement architecture is also one kind of deep generative model derived from GANs~\cite{NIPS2014_5423}. GANs work in a simple and efficient way to train both the generator and discriminator via the min-max two-player game, achieving remarkable results on unsupervised learning tasks.
Recently, GANs and their variants have shown great success in realistic image generation including super-resolution \cite{8099502}, image translation \cite{8237506}, and image synthesis \cite{8237891}. 
We here briefly review the recent applications of GANs on image generation.

Plenty of modified architectures of GANs~\cite{Wang_2018_CVPR,Bulat_2018_CVPR,Kossaifi_2018_CVPR,Li_2017_CVPR,Yi_2019_CVPR,Karras_2019_CVPR,Bao_2018_CVPR,Karras_2019_CVPR,Yin_2017_ICCV} are proposed to fulfill face generation.
For the face age progression problem, Yang et al. propose a pyramid architecture of GANs to learn face age progression, achieving remarkable results~\cite{Yang_2018_CVPR}. Besides, the identity-preserved conditional generative adversarial networks are proposed to generate a face with target age while preserving the identity~\cite{Wang_2018_CVPR}.

In the research area of face synthesis, Song et al. propose the geometry-guided GANs to achieve facial expression removal and synthesis~\cite{Song:2018:GGA:3240508.3240612}. Shen et al. synthesize faces while preserving identity with proposed three-player GAN called FaceID-GAN~\cite{Shen_2018_CVPR}. Bao et al. propose a framework based on GANs to disentangle identity and attributes of faces~\cite{Bao_2018_CVPR}, and recombine different identities and attributes for identity preserving face synthesis in open domain.  Li et al. propose an effective object completion algorithm using a deep generative model~\cite{Li_2017_CVPR}. Cao et al. propose CariGANs for unpaired photo-to-caricature translation which generates the caricature image in two steps of geometry and appearance adjustment~\cite{Cao:2018:CUP:3272127.3275046}.

Inspired by these works, we propose our facial attractiveness enhancement framework based on GANs to generate the enhanced and identity preserved facial images automatically. To the best of our knowledge, till now none of existing works based on GANs are specifically devised to fulfill this task. 

%% file: files/Method.tex
\section{Our Method}
\label{Methods}
\input{files/figure_overview.tex}
The original GANs consist of a generator $G$ and a discriminator $D$. $G$ and $D$ are trained alternatively and iteratively via an adversarial process. $G$ and $D$ compete with each other via a min-max game formulated as the following Equation,
\begin{equation}
\label{eq:classic_GAN}
\begin{aligned}
    \mathcal{V}_{D, G}= & \min \limits_{G} \max \limits_{D} \mathbb{E}_{x \sim P_{data}(x)}\log[D(x)] \\
    & + \mathbb{E}_{z \sim P_{z}(z)}\log[1 - D(G(z))],
\end{aligned}
\end{equation}
where $z$ represents the noise sampled from a prior probability distribution $P_z$, and $x$ denotes the data item sampled from real data distribution $P_{data}$. We aim to get $G$ until $D$ cannot distinguish real samples from the generated ones.

Let X and Y denote the domains of unattractive and attractive faces respectively. No paired faces exist between these two domains. Given an unattractive sample $x \in X$, our goal is to learn a mapping $\phi:X \to Y$ which can transfer $x$ to an attractive sample $y \in Y$.
Unlike previous works enhancing the facial attractiveness in either geometry structure or appearance texture, FA-GANs make the enhancement involving both of these two aspects. Therefore, FA-GANs consist of two parallel branches: the geometry adjustment network $\phi_{geo}$ and the appearance adjustment network $\phi_{app}$.

The proposed framework of FA-GANs with two branches is shown in Fig.~\ref{fig:flowchart}. Each branch basically is with the architecture of GANs. To enhance face attractiveness in both the geometry and appearance aspects, the geometry branch is trained to learn the geometry-to-geometry translation from $X$ to $Y$. The appearance branch is trained with the help of geometry translation $\phi_{geo}$ to learn the adjustment in both of the appearance and geometry aspects.
Similar to the traditional beautification engine~\cite{Leyvand:2008:DEF:1399504.1360637} of 2D facial images, facial landmarks are used to represent the geometry structure. We denote $L_X$ and $L_Y$ as the domains of geometry structures of unattractive and attractive faces, respectively.
The geometry branch learns the mapping $\phi_{geo}: L_X \to L_Y$ to convert geometry structure of the unattractive face $l_x \in L_X$ to the attractive one $l_y \in L_Y$. The appearance branch learns to convert the instance $x$ to an intermediate instance $y’ \in Y’$, where $y’$ is an appearance adjusted face with the same geometry structure as $x$, and $Y’$ is the intermediate domain with geometry structure of $X$ and appearance texture of $Y$. The mapping of appearance branch learned is defined as $\phi_{app}: X \to Y’$.

To combine these two branches and enhance the facial attractiveness in both aspects, we further minimize the energy function to ensure the consistency between the geometry structure $l_{y'}$ of the intermediate face $y'$ and the geometry structure $G_{geo}(x)$ generated by geometry branch. The consistency energy is defined as:
\begin{equation}
\begin{aligned}
    \mathbf{E}_{con} = \mathbb{E}_{x \in X}\Vert G_{geo}(l_x) - l_{y'} \Vert_2.
\end{aligned}
\end{equation}
To achieve this, we further introduce the pre-trained geometry analysis networks to extract the geometry structure $l_{y’}$ from $y’$ generated by the appearance branch.
FA-GANs combine the geometry mapping $\phi_{geo}: L_X \to L_Y$ and appearance mapping $\phi_{app}: X \to Y’$ and achieve the mapping $\phi:X \to Y$ by enforceing the consistency between the geometry and appearance branches.
The geometry branch, appearance branch, and the combination of these two branches are  described in detail in the following.

\subsection{Geometry Enhancement Branch}
\label{section:FacialGeometryEnhancementBranch}
\input{files/figure_overGeoEnhance.tex}
\subsubsection{Geometry data}
We extract 2D facial landmarks for each of the training face images. For each face, $60$ landmarks are detected, and the detected landmarks are located on the outlines of left eyebrow, right eyebrow, left eye, right eye, nose bridge, nose tip, top lip, bottom lip, and chin. Fig.~\ref{fig:overviewGeoEnhance} shows an example of extracted landmarks. To normalize the scale, all faces are cropped and resized to the size of $224 \times 224$ pixels. Instead of learning from facial landmarks directly, we prefer to enhance facial attractiveness via adjusting the relative distances among important facial components, such as eyes, nose, lip, and chin. Landmarks are more sensitive to small errors than relative distances. Therefore, the detected landmarks are used to construct a distance embedded mesh containing $150$ edges through Delaunay triangulation. We further reduce the dimension of the distance vector to $32$ by applying principal component analysis (PCA) and reserve $99.01\%$ of total variants.

\subsubsection{Geometry enhancement networks}
Geometry branch aims to learn the mapping $\phi_{geo}: L_X \to L_Y$, while no samples in $L_X$ and $L_Y$ are paired. We achieve this by constructing the geometry adjustment GANs including only full connected (FC) and ReLU~\cite{Nair:2010:RLU:3104322.3104425} layers. Unlike traditional GANs which train the generator until it fools the discriminator, the feature maps of $l_y \in L_Y$ are also utilized to help to learn the geometry mapping. The generator takes $l_x \in R^{32}$ as input and outputs the adjusted distance vector $l_y \in R^{32}$. Fig.~\ref{fig:overviewGeoEnhance} shows the architecture of our geometry adjustment branch. Apart from the last layer, the discriminator has the same architecture with the generator. The details of the architectures are shown in Table~\ref{tab:branchOfGAGANs}.
\begin{table}
\caption{Architecture of geometry enhancement branch.}
\centering
\begin{tabular}[p]{c c c}
\hline
Module & Layer & Activation size \\ \hline
& Input & $32$ \\
& FC-ReLU & $64$\\
& FC-ReLU & $64$\\
& FC-ReLU & $64$\\
& FC-ReLU & $64$\\
\hline
Generator & FC & $32$\\
\hline
Discriminator & FC & 1\\
\hline
\end{tabular}
\label{tab:branchOfGAGANs}
\end{table}

Following the classic GANs~\cite{NIPS2014_5423, Mao_2017_ICCV}, the process of training discriminator amounts to minimizing the loss:
\begin{equation}
\label{eq:geometry_classicGAN_D0}
\begin{aligned}
    \mathcal{L}_{geo\_D}= & \mathbb{E}_{l_x \in L_X}[D_{geo}(G_{geo}(l_x))^2] \\
    +& \mathbb{E}_{l_y \in L_Y}[(1 - D_{geo}(l_y))^2],
\end{aligned}
\end{equation}
where $D_{geo}$ and $G_{geo}$ are the discriminator and the generator, respectively. $D_{geo}$ takes both the unattractive samples and generated samples as negative samples and regards the attractive samples as positive samples. The loss of discriminator can be rewritten as:
\begin{equation}
\label{eq:geometry_classicGAN_D0}
\begin{aligned}
    \mathcal{L}_{geo\_D}= & \mathbb{E}_{l_x \in L_X}[D_{geo}(G_{geo}(l_x))^2] \\
    + & \mathbb{E}_{l_x \in L_X}[D_{geo}(l_x)^2] \\
    + & \mathbb{E}_{l_y \in L_Y}[(1 - D_{geo}(l_y))^2].
\end{aligned}
\end{equation}
Considering that samples coming from the same category have similar feature maps, the feature loss is introduced and the loss of generator is defined as:
\begin{equation}
\label{eq:geometry_classicGAN_G}
\begin{aligned}
    \mathcal{L}_{geo\_G}=
    & \mathbb{E}_{l_x \in L_X}[( 1 - D_{geo}(G_{geo}(l_x)))^2] \\
    + & \mathbb{E}_{l_x \in L_X, l_y \in L_Y}\Vert fc_i(G_{geo}(l_x)) - fc_i(y)\Vert_2 \\
    + & \mathbb{E}_{l_x \in L_X, l_y \in L_Y}\Vert G_{geo}(l_x) - l_y\Vert_2,
\end{aligned}
\end{equation}
where $fc$ is the FC layer. $i \in \{1, 2, 3, 4\}$  indicates the i-th layer in the generator $G_{geo}$.
\subsubsection{Geometry adjustment on faces}
\label{geometry_warpping}
Geometry branch converts $l_x$ to $l_y$ with $l_x$ being the distance vector derived from the landmark points $\mathbf{p}_x$.
To adjust facial geometry, the enhanced points $\mathbf{p}_y$ corresponding to $l_y$ are estimated by minimizing the following energy function for the best fit,
\begin{equation}
\begin{aligned}
\label{eq:energy}
    E({p_y}_1,...,{p_y}_n)=\sum_{e_{ij}}{({\Vert {p_y}_i - {p_y}_j \Vert}_2 - d_{ij}^2)}^2,
\end{aligned}
\end{equation}
where $e_{ij}$ is the element of the connectivity matrix of our constructed facial mesh. The distance term $d_{ij}$ is the entry in $l_y$ corresponding to $e_{ij}$.
Minimization of the energy function is performed by the Levenberg-Marquardt~\cite{more1978levenberg} algorithm.
Then the geometry enhanced face is generated by mapping the original face texture from the mesh constructed by $\mathbf{p}_x$ to the mesh of $\mathbf{p}_y$.

\subsection{Appearance Enhancement Branch}
\label{section:FacialAppearanceEnhancementBranch}

The outstanding performance of GANs in fitting data distribution has significantly promoted many computer graphics and vision applications such as image-to-image translation~\cite{8100115, 8237506, Cao:2018:CUP:3272127.3275046, Li:2018:BIF:3240508.3240618}. Inspired by these studies, we employ GANs to perform facial appearance enhancement while preserving identity. The appearance adjustment networks and the loss are introduced in the following.

\subsubsection{Appearance adjustment networks}
The process of appearance enhancement only requires a forward pass through the generator $G_{app}$. The generator $G_{app}$ is designed with the U-Net~\cite{10.1007/978-3-319-24574-4_28}. The discriminator $D_{app}$ is introduced to output the indicators $D_{app}(x)$ suggesting the probability that $x$ comes from the attractiveness category. Different from the classic discriminator of GANs, we implement our discriminator with the pyramid architecture~\cite{Yang_2018_CVPR} to estimate high-level attractiveness-related cues in a fine-grained way.
	
Specifically, a facial attractiveness classifier is pre-trained with the architecture of VGG-16~\cite{DBLP:journals/corr/SimonyanZ14a} to classify attractive and unattractive faces. The hierarchical layers of VGG-16 endow our network with the ability to capture image features from the pixel level to semantic level. The generator $G_{app}$ is optimized until the discriminator $D_{app}$ is confused about $G_{app}(x)$ and $y \in Y$ for all the pixel and semantic level features. Consequently, the feature maps of the 2nd, 4th, 7th, and 10th convolutional layers in VGG-16 are integrated into the discriminator $D_{app}$ for adversarial learning.
The generator $G_{app}$ not only transfers $x \in X$ to $y \in Y$ but also preserves the identity. We achieve this by leveraging the network of deep face descriptor~\cite{BMVC2015_41} $\psi_{id}$ to measure the identity similarity between $x$ and $G_{app}(x)$. The deep face descriptor $\psi_{id}$ is trained on a large face dataset containing millions of facial images by recognizing $N=2622$ unique individuals. We remove the classification layer and take the last FC layer as the identity preserving output, forming an identity descriptor $\psi_{id}(x) \in R^{4096}$. Both the original face $x$ and enhanced face $G_{app}(x)$ are fed into $\psi_{id}$ to generate $G_{app}(x)$ with small margin between $\psi_{id}(G_{app}(x))$ and $\psi_{id}(x)$.
\input{files/tab_extractor.tex}

\subsubsection{Loss} The appearance improved face should be recognized as the same individual as the original one. To this end, the loss concerning  appearance enhancement should not only take effect in helping improve the facial attractiveness, but also facilitate the preservation of identity. In addition, the difference between the improved face and the input should be constrained. Considering all these aspects, four types of loss for training the appearance branch are defined. They are the adversarial loss, identity loss, pixel loss, and total variation loss.

a) Adversarial loss. Similar to the geometry adjustment branch, both unattractive faces and the generated faces are deemed as negative faces and the attractive faces are deemed as positive ones. We also adopt the adversarial loss of LSGAN~\cite{Mao_2017_ICCV} to train the appearance adjustment branch. The adversarial loss is defined as:
\begin{equation}
\label{eq:appearance_classicGAN_D}
\begin{aligned}
    \mathcal{L}_{app\_D}= & \mathbb{E}_{x \in X}[D_{app}(G_{app}(x))^2] \\
    + & \mathbb{E}_{x \in X}[D_{app}(x)^2] \\
    + & \mathbb{E}_{y \in Y}[( 1 - D_{app}(y))^2],
\end{aligned}
\end{equation}
\begin{equation}
\label{eq:appearance_classicGAN_G}
\begin{aligned}
    \mathcal{L}_{app\_G}= & \mathbb{E}_{x \in X}[(1-D_{app}(G_{app}(x)))^2]. \\
\end{aligned}
\end{equation}

b) Identity loss. To ensure that appearance enhanced face $G_{app}(x)$ and the original face $x$ are recognized as the same individual, the identity loss is introduced and is defined as:
\begin{equation}
\begin{aligned}
    \mathcal{L}_{id} = \mathbb{E}_{x \in X}\Vert \psi_{id}(x) - \psi_{id}(G_{app}(x)) \Vert_2,
\end{aligned}
\end{equation}
where $\psi_{id}$ is the identity descriptor derived from the deep face descriptor~\cite{BMVC2015_41}.

c) Pixel loss. The generated face $G_{app}(x)$ is more attractive than the input one $x$. However, the gap between $G_{app}(x)$ and $x$ in pixels should be constrained. The pixel loss $\mathcal{L}_{px}$ enforces the enhanced face $G_{app}(x)$ to have small difference with the original face $x$ in the raw-pixel space. In addition, in experiences of other image generation tasks, $L1$ regularization performs better with less blurring than $L2$ regularization. The pixel loss is thus formulated as:
\begin{equation}
\begin{aligned}
    \mathcal{L}_{px} = \frac{1}{W \times H \times C} \sum \vert x - G_{app}(x) \vert,
\end{aligned}
\end{equation}
where $W$ and $H$ are the width and height of the image, respectively, and $C$ is the number of channels.

d) Total variation loss. Total variation loss $\mathcal{L}_{tv}$ is defined as total variation regularizer~\cite{1510697} in order to encourage spatial smoothness of the enhanced face area $G_{app}(x)$.

\subsection{Geometry Enhancement Consistency}
\label{section:geometryEnhancementConsistency}
Given a face $x$ as well as its distance vector $l_x$, geometry enhancement branch converts $l_x$ to $G_{geo}(l_x)$, where $G_{geo}(l_x)$ lies in the domain of $L_Y$. At the same time, appearance branch converts $x$ to $G_{app}(x)$, where $G_{app}(x)$ lies in the domain of $Y$. As illustrated in Fig.~\ref{fig:flowchart}, FA-GANs combine these two branches and output the enhanced face with the generator of the appearance enhancement branch. The geometry consistency exists between $G_{geo}(l_x)$ and $l_{G_{app}(x)}$, where $l_{G_{app}(x)}$ is the distance vector of $G_{app}(x)$. To extract the distance vector, an extractor $E$ is pre-trained on our dataset. The extractor is constructed with convolutional and FC layers, and each convolutional layer is followed by a Batch Normalization~\cite{Ioffe:2015:BNA:3045118.3045167} layer and a ReLU~\cite{Nair:2010:RLU:3104322.3104425} layer. The detailed architecture of the extractor is shown in Table~\ref{tab:extractor}.

The extracted vector $E(x)$ is the PCA representation of distance vectors derived from landmarks. Thus, the geometry enhancement consistency is defined with the following loss:
\begin{equation}
\begin{aligned}
    \mathcal{L}_{gec} = \mathbb{E}_{x \in X}\Vert G_{geo}(l_x) - E(G_{app}(x)) \Vert_2.
\end{aligned}
\end{equation}

In summary, FA-GANs contain the losses in the appearance branch and the loss enforcing geometry enhancement consistency. Therefore, the overall training loss is expressed as,
\begin{equation}
\label{eq:FA-GANs}
\begin{aligned}
    \mathcal{L}_{G} =
    &\lambda_1 \mathcal{L}_{app\_G} + \lambda_2 \mathcal{L}_{id} + \lambda_3 \mathcal{L}_{px}\\
    +&\lambda_4 \mathcal{L}_{tv} + \lambda_5 \mathcal{L}_{gec},
\end{aligned}
\end{equation}
\begin{equation}
\begin{aligned}
    \mathcal{L}_{D} = \mathcal{L}_{app\_D}.
\end{aligned}
\end{equation}
We train $G$ and $D$ alternately until $G$ learns the desired facial attractiveness transformation and $D$ becomes a reliable estimator.

%% file: files/figure_overview.tex
\begin{figure*}[tb]
\centering
\includegraphics[width=1.0\linewidth]{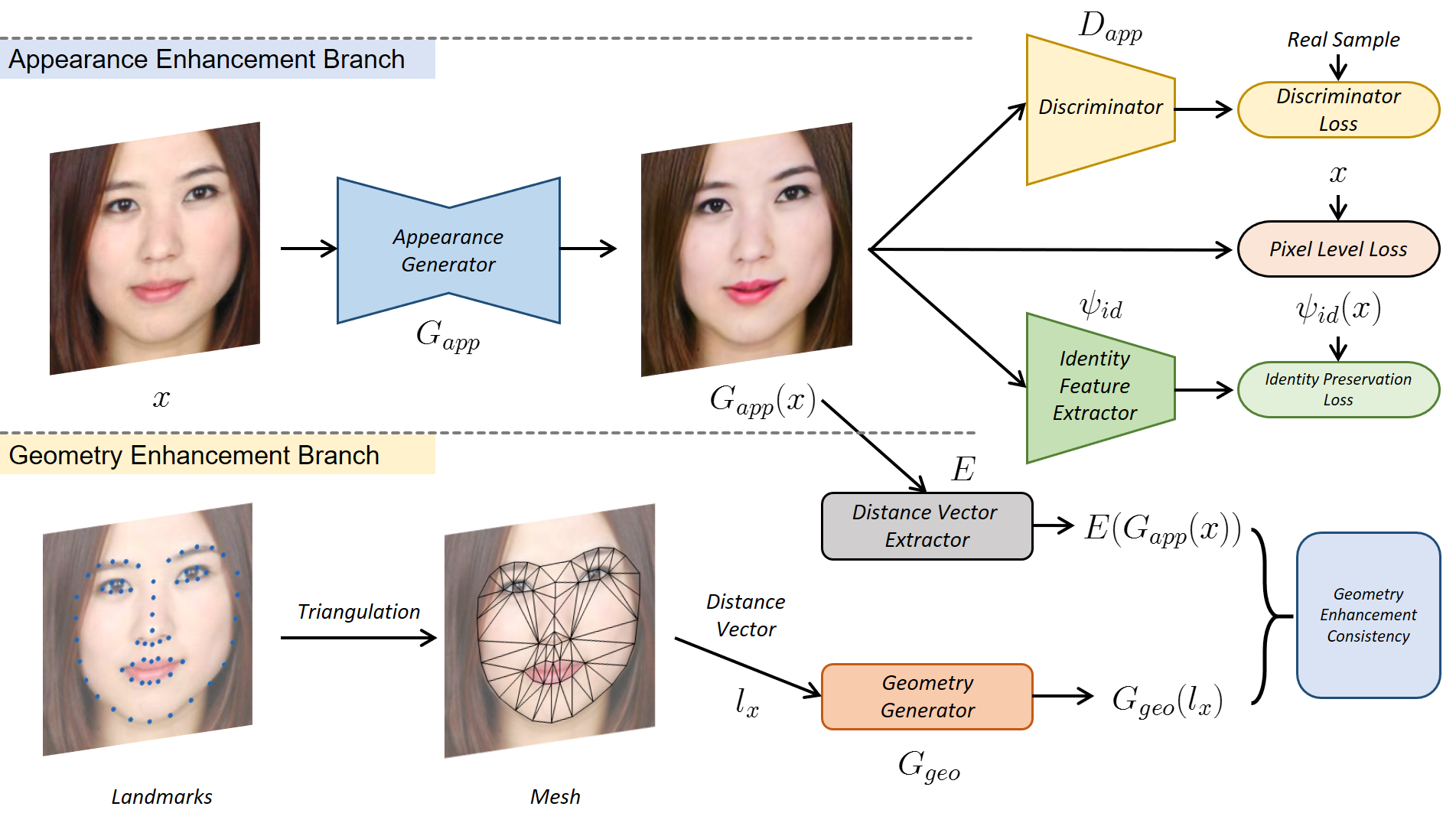}
\caption{The systematic overview of FA-GANs. FA-GANs consist of two branches, and they are appearance enhancement branch and geometry enhancement branch. Geometry enhancement consistency between these two branches is used for learning the geometry adjustment in appearance generator. }	
\label{fig:flowchart}	
\end{figure*} 

%% file: files/figure_overGeoEnhance.tex
\begin{figure*}[tb]
\centering
\includegraphics[width=\linewidth]{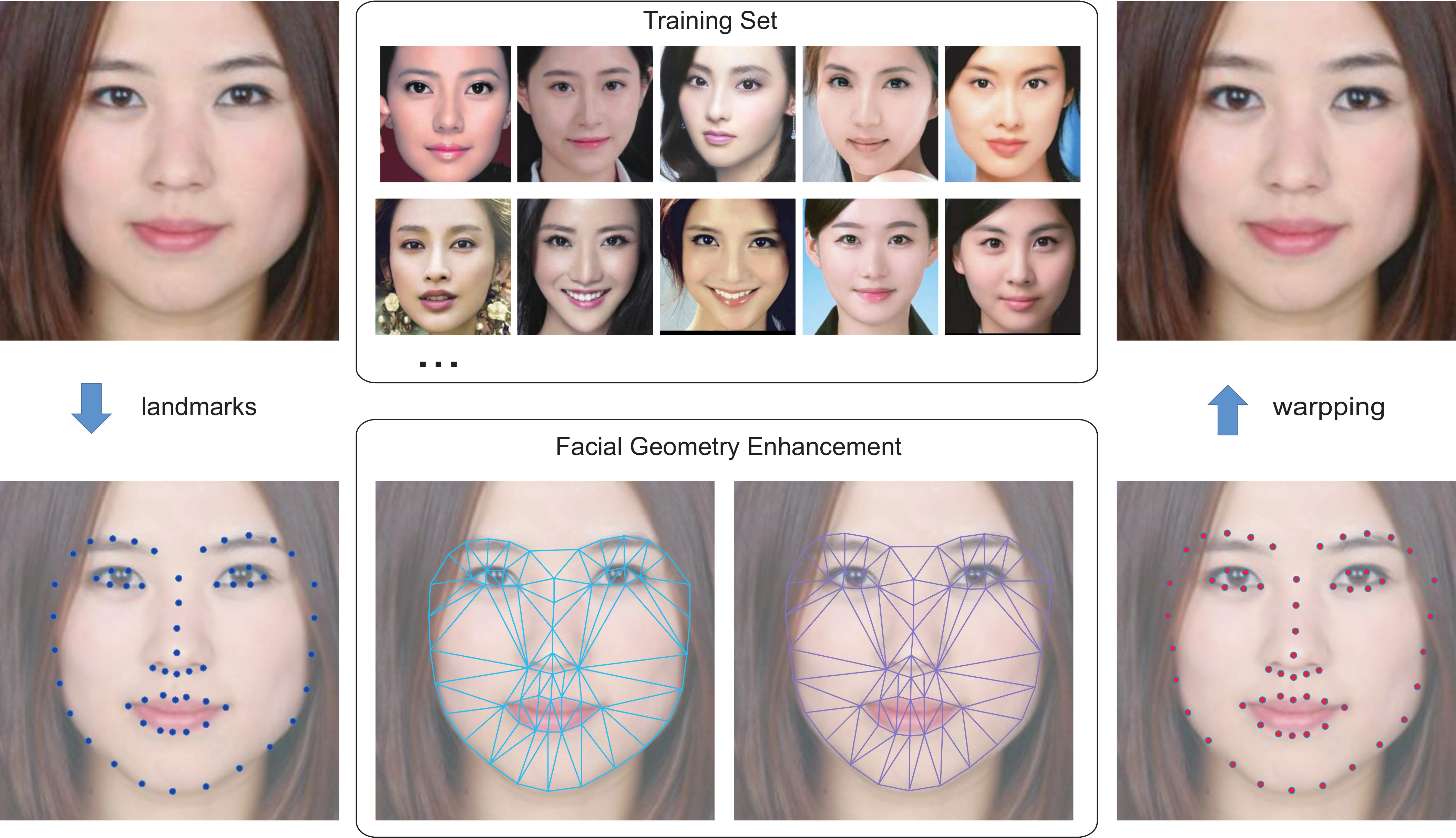}
\caption{Framework of our facial geometry enhancement. Geometry enhancement is performed on the domain of distance vectors, which are derived from the landmarks detected on the original face. And then, the enhanced landmarks are inferred from the enhanced distance vectors. At last, the geometry enhanced face is generated by mapping the original face to the enhanced landmarks.}	
\label{fig:overviewGeoEnhance}	
\end{figure*} 

%% file: files/tab_extractor.tex
\begin{table}
\centering
\caption{Convolutional layers of the distance feature extractor.}
\label{tab:extractor}
\begin{tabular}[p]{c c}
\hline
Layer & Activation size\\ \hline
Input & 3$\times$224$\times$224\\
$32 \times 9 \times 9$ stride $1$ padding $4$ &	$32 \times 224 \times 224$ \\
$64 \times 3 \times 3$ stride $2$ padding $1$	& $64 \times 112 \times 112$ \\
$128 \times 3 \times 3$ stride $2$ padding $1$ &	$128 \times 56 \times 56$ \\
$128 \times 3 \times 3$ stride $1$ padding $1$ &	$128 \times 56 \times 56$ \\
$128 \times 3 \times 3$ stride $1$ padding $1$ &	$128 \times 56 \times 56$ \\
$128 \times 3 \times 3$ stride $1$ padding $1$ &	$128 \times 56 \times 56$ \\
$64 \times 3 \times 3$ stride $2$ padding $1$	& $64 \times 28 \times 28$ \\
$32 \times 3 \times 3$ stride $2$ padding $1$	& $32 \times 14 \times 14$ \\
$16 \times 3 \times 3$ stride $2$ padding $1$	& $16 \times 7 \times 7$ \\
\hline
FC-ReLU & $784$ \\
FC-ReLU & $512$ \\
FC & 32 \\
\hline
\end{tabular}
\end{table}

%% file: files/Experiment.tex
\input{files/fig_branch_geo.tex}
\section{Experiments}
In this section, we first report the face dataset used in our experiments. We then analyze the performance of the appearance and geometry branches of our FA-GANs, which corresponds to the ablation study of geometry enhancement consistency module. We finally conduct comprehensive experiments to compare FA-GANs with the state-of-the-arts, including the automatic function of facial enhancement provided by prevalent mobile APPs such as Meitu~\cite{Meitu}. In addition to the results generated by FA-GANs, we also show the results obtained by applying our geometry branch and appearance branch successively to the testing faces and compare with relevant methods. 
\label{Experiments}
\subsection{Datasets}
Facial attractiveness analysis is fundamental to facial image enhancement. The attractiveness criterion is explored in a data-driven way in our paper. However, few datasets designed specifically for facial attractiveness analysis are publicly available. Fortunately, Liang et al. propose a diverse benchmark dataset, called SCUT-FBP5500, for the prediction of multi-paradigm facial beauty~\cite{liang2017SCUT}. The SCUT-FBP5500 dataset contains totally 5500 frontal faces involving diverse properties, such as male/female, Asian/Caucasian, and ages. Specifically, it includes 2000 Asian females, 2000 Asian males, 750 Caucasian females, and 750 Caucasian males, mostly aged from 15 to 60 with neutral expression. The attractiveness score of each facial image is labeled with 60 different people on a five-point holistic scale, where 1 indicates the most unattractive and 5 means the most attractive. Our attractiveness learner classifies the faces into two categories of attractiveness and unattractiveness. In order to obtain a dataset with unanimous labels, the faces that have conflict scores are ruled out. Moreover, we enlarge the dataset by further collecting the portraits of famous beautiful actresses as attractive faces and the portraits of some selected ordinary people as unattractive faces. Totally, $7798$ female facial images are collected for FA-GANs. The attractiveness and unattractiveness categories contain $3702$ and $4096$ images, separately.

\input{files/fig_branch_app.tex}

\input{files/fig_faceParsing.tex}

\input{files/fig_expRes.tex}

\subsection{Implementation Details}
All faces are detected and their landmark points are extracted. The geometric information extracted are fed to the geometry enhancement branch. All faces are cropped and scaled to the size of $224 \times 224$ pixels as the training data of our appearance enhancement branch. To get a more robust attractiveness ranking module, we also add the color jitter, random affine, random rotation, and random horizontal flip to the facial attractiveness network.
The trade-off parameters $\lambda_1$, $\lambda_2$, $\lambda_3$, $\lambda_4$, and $\lambda_5$ in Equation~\eqref{eq:FA-GANs} are set to $10.0$, $1.0$, $5.0$, $10^{-5}$, and $10^{3}$, separately. The Adam~\cite{kingma2014adam} algorithm with the learning rate of $1\times 10^{-4}$ is used to optimize FA-GANs. FA-GANs are trained on a computer with a 4.20GHz 4-core, i7-7700K Intel CPU and a GTX 1080Ti GPU, which costs around $5$ hours for $15000$ iterations with the batch size of $8$ to generate the desired results.

\subsection{Branch Analysis of FA-GANs}
To investigate the performance of the appearance branch and geometry branch of our FA-GANs, we perform ablation studies and analyze the effect of each branch. These two branches are the variations of GANs. In implementation, the geometry branch is pre-trained for training FA-GANs. Hence, we explore the performance of appearance branch by training it without the geometry consistency constraint.
The adjustment results of geometry and appearance branches are demonstrated in Figs.~\ref{fig:branchgeo} and~\ref{fig:branchapp}.
As seen from Fig.~\ref{fig:branchgeo}, geometry branch enhances the facial attractiveness by generating faces with small cheeks and big eyes. This suggests that people prefer smaller faces and bigger eyes in nowadays. Besides, appearance branch adjusts faces mainly on the texture instead of the geometry structure as shown in Fig.~\ref{fig:branchapp}. It tends to generate faces with clean cheeks, red lips, and black eyes, which reveals the popular cognition of beauty. We further analyze the results of attractiveness in qualitative evaluation and time consuming in the following subsection.

\input{files/fig_statistical.tex}

\subsection{Comparison with State-of-the-Arts}
We verify the effectiveness of FA-GANs by comparing our generated results with the results by existing methods in geometry and appearance adjustment. Specifically, we make comparisons with the geometry adjustment method proposed in~\cite{Leyvand:2008:DEF:1399504.1360637} and the appearance adjustment method with photo-realistic image style transfer~\cite{Li_2018_ECCV}.
We implement~\cite{Leyvand:2008:DEF:1399504.1360637} with its KNN-based beautification engine. $K$ beautiful faces are searched in the domain of $L_Y$, and $K$ is set to 3.The geometry structure is then adjusted as described in subsection~\ref{geometry_warpping}. Photorealistic image style transfer~\cite{Li_2018_ECCV} needs a face $r\in Y$ as the reference in order to transfer its style to the input face $x\in X$.
We choose $r$ by selecting the nearest face in the domain of $L_Y$, which can be obtained in the implementation of~\cite{Leyvand:2008:DEF:1399504.1360637} with $K=1$. With the help of facial landmarks we detected, the input face $x$ and the reference face $r$ are further parsed\footnote{https://github.com/zllrunning/face-parsing.PyTorch} for~\cite{Li_2018_ECCV}, and the exemplar parsed faces are shown in Fig.~\ref{fig:faceParsing}.
Furthermore, the attractiveness enhancement results involving these two aspects are also analyzed. Given a face, we first adjust its geometry structure with~\cite{Leyvand:2008:DEF:1399504.1360637} or our geometry branch and then adjust its appearance with our appearance branch or~\cite{Li_2018_ECCV}. Moreover, we also compare FA-GANs to BeautyGAN~\cite{Li:2018:BIF:3240508.3240618} with the reference face selected for ~\cite{Li_2018_ECCV}.

\input{files/tab_identity.tex}

\input{files/tab_time.tex}

Some mobile APPs are specifically designed for beautifying facial images and the automatic beautification on facial images is supported. We also make a comparison with the APP of Meitu, which is famous for beautifying the faces. Each face is adjusted by the “One-click beauty” in face beautification module of Meitu and with the default settings.

\subsubsection{Qualitative evaluation}
The comparison results are shown in Fig.~\ref{fig:expRes}, and are arranged into six groups. They are original, geometry adjustment, appearance adjustment, two-step adjustment, the state-of-the-arts, and FA-GANs, respectively.

In Fig.~\ref{fig:expRes}, Leyvand et al.~\cite{Leyvand:2008:DEF:1399504.1360637} adjusts the geometry structure effectively compared to the original faces, and it also tends to generate portrait with smaller cheeks and bigger eyes.
Li et al.~\cite{Li_2018_ECCV} adjust the appearance mainly depending on the reference image, and the generated face looks homogeneous on the face area.
BeautyGAN~\cite{Li:2018:BIF:3240508.3240618} only adjusts the facial image in the appearance aspect, and it has no effects in the geometry aspect. Moreover, BeautyGAN adjusts the facial image depends on the reference image and transfer the makeup style to the non-makeup image.
Compared to BeautyGAN, ``One-click beauty” of Meitu adjusts the face in both aspects. All of the instances generated by Meitu shrink the cheek slightly and have whiter skins while the colors of mouth and eyebrows are diluting. Our appearance branch promotes the original faces greatly and also generates whiter skin and has the makeup style on the mouth, eyes, and eyebrows. The geometry branch adjusts the geometry structure and tends to generate smaller cheeks and bigger eyes in a data-driven manner. FA-GANs adjust the face in both appearance and geometry aspects. The enhanced faces have whiter skin and the makeup style as appearance branch and have adjusted geometry structure as geometry branch.

\subsubsection{Quantitive evaluation}
To assess the results by different methods quantitatively, the facial attractiveness is assessed by Face++ API\footnote{https://www.faceplusplus.com/beauty/}, which provides users with accurate, efficient, and reliable face-based services.
It gives two scores: $f_1\%$ and $f_2\%$ indicating that males and females generally think this face is more attractive than $f_1\%$ and $f_2\%$ persons, respectively.
$100$ groups of the original faces and their results are evaluated. The averaged attractiveness scores for each method are shown in Fig.~\ref{fig:statistical}.

All these methods get higher attractiveness scores than the original faces. The statistical results suggest that FA-GANs achieve the best performance over all these methods. The two-step adjustment method of applying our geometry branch followed by appearance branch to the input faces achieves the second best.
FA-GANs promote the attractiveness score from $62.21\%$ to $70.23\%$ and $61.44\%$ to $70.70\%$ in the female and male views, respectively. It suggests that FA-GANs enhance the attractiveness effectively.
Similarly, the two-step adjustment method achieves the second best, getting scores of $68.76\%$, $68.23\%$, respectively and outperforms BeautyGAN ($67.72\%$, $67.28\%$) and Meitu ($66.77\%$, $65.94\%$).
Comparing the geometry adjustment results of our geometry branch in FA-GANs and~\cite{Leyvand:2008:DEF:1399504.1360637}, geometry branch outperforms~\cite{Leyvand:2008:DEF:1399504.1360637} with a small margin in both assessing aspects of female and male. On the other hand, the appearance branch performs better than~\cite{Li_2018_ECCV} no matter directly applying these appearance adjusting methods directly to the original faces or to the intermediate results obtained by other geometry adjusting methods.
Comparing the results between appearance and geometry adjustment, appearance always achieves higher scores than geometry indicating that people can enhance their attractiveness by paying more attention to makeups than facelifts.

For identity preservation, we measure it using the cosine distance between $\psi_{id}(x)$ and $\psi_{id}(x’)$ where $x$ and $x’$ are the original face and the adjusted face, respectively, and $\psi_{id}$ is the deep face descriptor. The evaluated similarity scores are shown in Table~\ref{tab:identity} with respect to the listed methods in Fig.~\ref{fig:statistical}.
As can be seen, all of the methods preserve the identity with the similarity greater than $80.00\%$. Meitu preserves the identity best with the similarity of $97.73\%$. Our geometry branch achieves the second best with the similarity of $96.34\%$, and FA-GANs also performs well with the similarity of $93.91\%$.

\subsubsection{Runtime time analysis}
To build an effective and efficient method for enhancing the attractiveness of faces, we further compare these methods in time consuming. All the time consuming experiments are performed on the aforementioned computer. Geometry adjustment can be recognized as three steps, which contain extracting distance vector $l_x$ from original face $x$, enhancing $l_x$ to $l_y$, and mapping $l_y$ to the enhanced face $y$. The time consuming analysis of these steps is shown in Table~\ref{tab:geoAdjust}.  As can be seen in Table~\ref{tab:geoAdjust},  Leyvand et al.~\cite{Leyvand:2008:DEF:1399504.1360637} performs faster than our geometry branch in $0.1832$ seconds on our geometry adjustment dataset. The geometry adjustment methods spend most time on mapping $l_y$ to enhanced face $y$.
Appearance branch enhances the attractiveness with only a forward pass through the generator. However, Li et al.~\cite{Li_2018_ECCV} adjusts the appearance with an extra reference image, and requires image parsing in order to get a better result. The time consumption of appearance adjustment is shown in Table~\ref{tab:appAdjust}. It suggests that appearance branch only requires $0.0599$ seconds to make adjustment on average. At last, we further compare the time consuming in the two-step adjustment method and FA-GANs in Table~\ref{tab:twoStepAdjust}. It demonstrates that FA-GANs is the fastest. An input face can be adjusted in only $0.0552$ seconds. Furthermore, we compare FA-GANs with BeautyGAN in adjusting an input facial image. FA-GANs also performs faster than BeautyGAN, which takes $0.0794$ seconds on average.

%% file: files/fig_branch_geo.tex
\begin{figure}[tb]
\centering
\includegraphics[width=\linewidth]{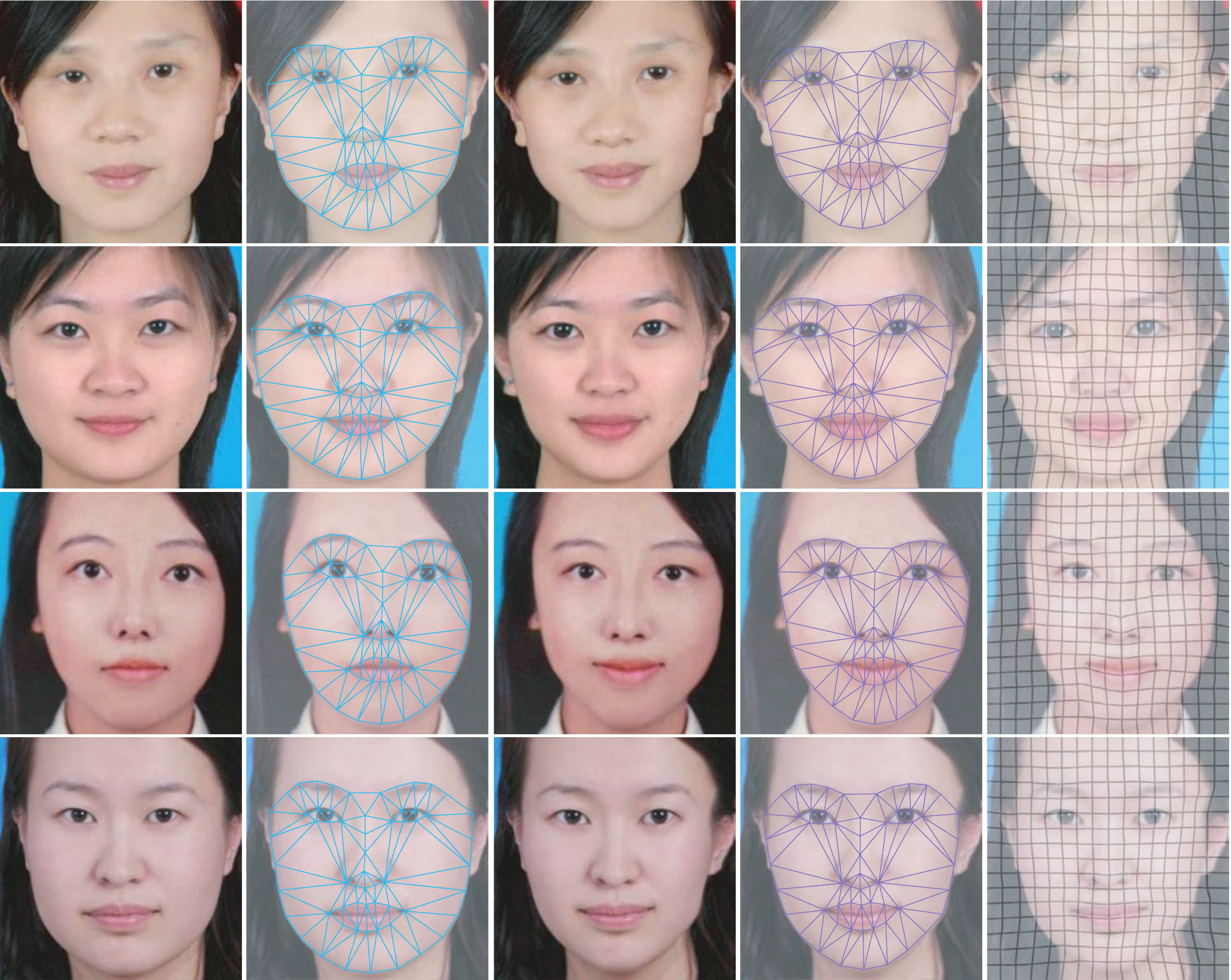}
\caption{Results of geometry branch adjustment. The original face, distance vectors of original face, adjusted face, distance vectors of adjusted face, and deformation grid are displayed from left to right in each row, respectively.}	
\label{fig:branchgeo}	
\end{figure}

%% file: files/fig_branch_app.tex
\begin{figure}[tb]
\centering
\includegraphics[width=\linewidth]{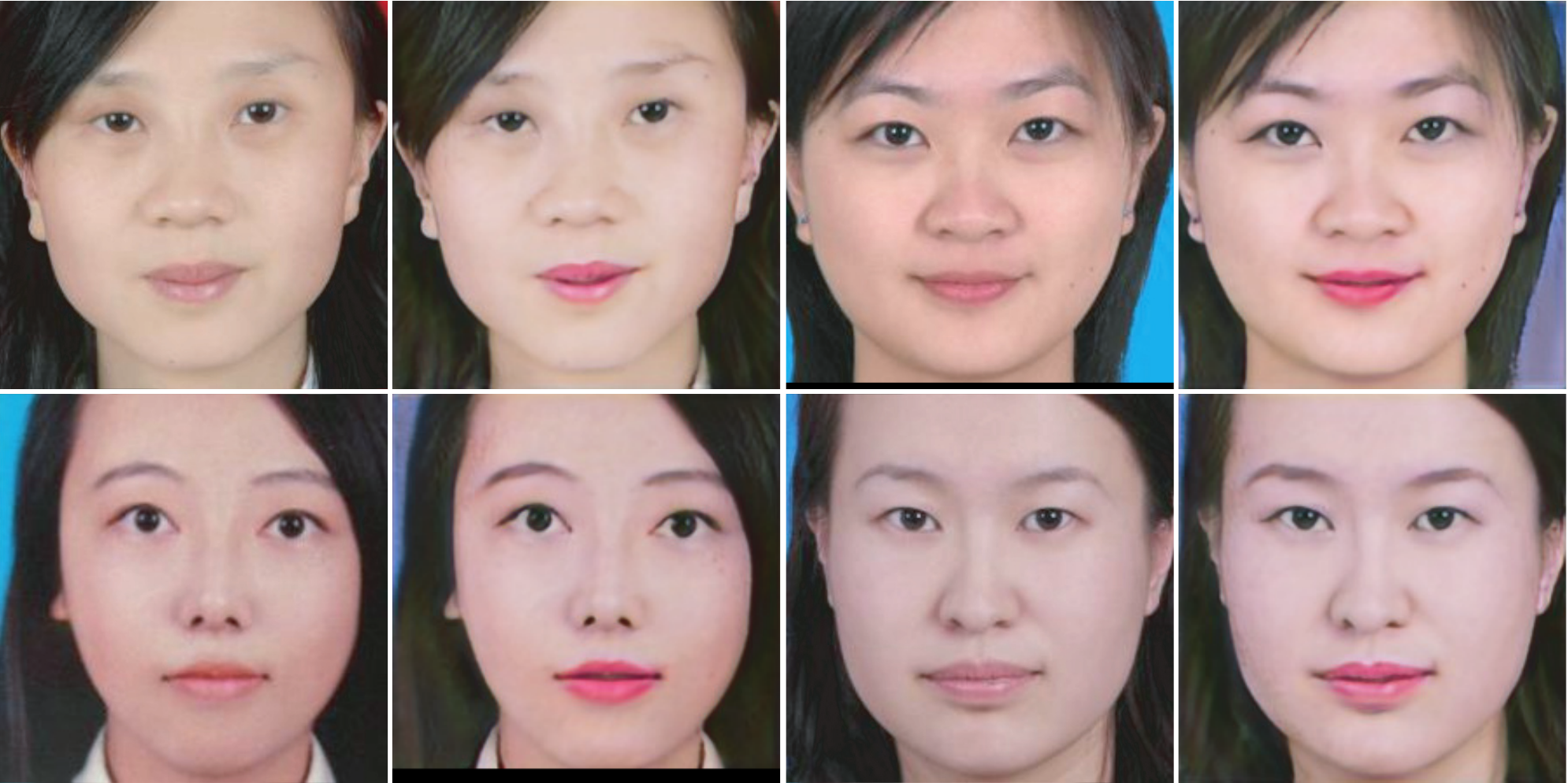}
\caption{Results of appearance branch adjustment. Each pair contains the input face and the appearance adjusted face.}
\label{fig:branchapp}	
\end{figure}

%% file: files/fig_faceParsing.tex
\begin{figure}[tb]
\centering
\includegraphics[width=\linewidth]{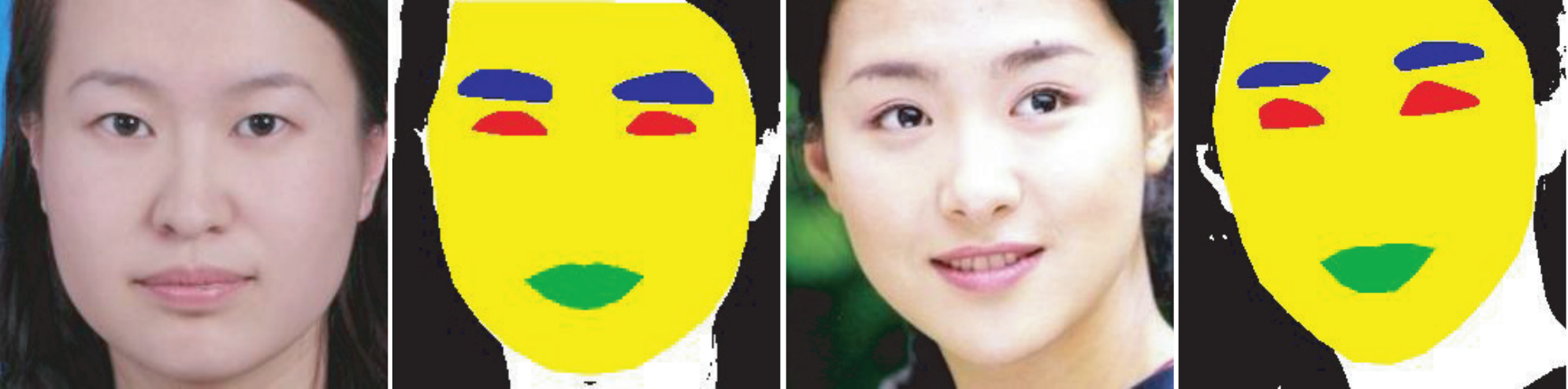}
\caption{Parsed faces used for ~\cite{Li_2018_ECCV}. The original as well as its parsed face and the reference as well as its parsed face are shown on left and right, respectively. }	
\label{fig:faceParsing}	
\end{figure}

%% file: files/fig_expRes.tex
\begin{figure*}[tb]
\centering
\includegraphics[width=\linewidth]{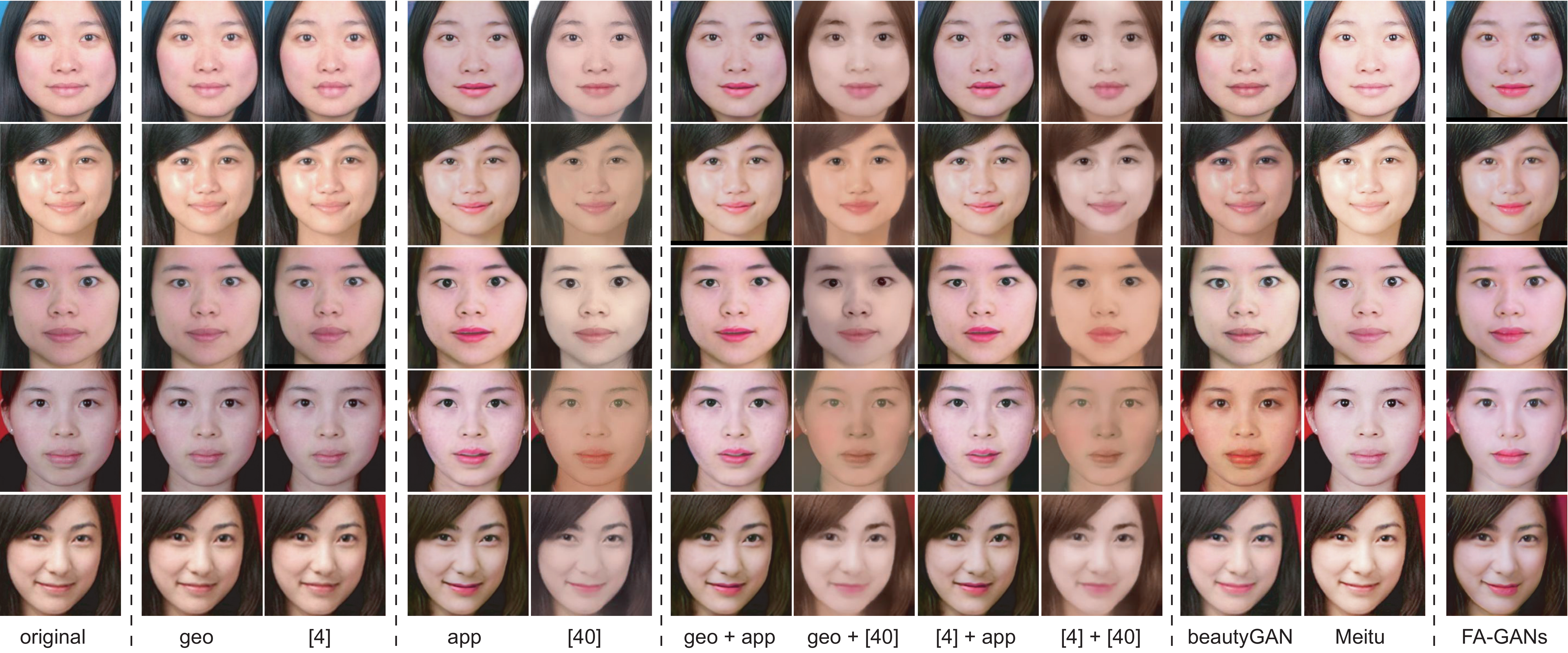}
\caption{Exemplar photos of original, geometry adjusted, appearance adjusted, two-step adjusted, and FA-GANs. Both the geometry and appearance aspects contain the results of branch of FA-GANs and the compared methods.}	
\label{fig:expRes}	
\end{figure*}

%% file: files/fig_statistical.tex
\begin{figure*}[tb]
\centering
\includegraphics[width=\linewidth]{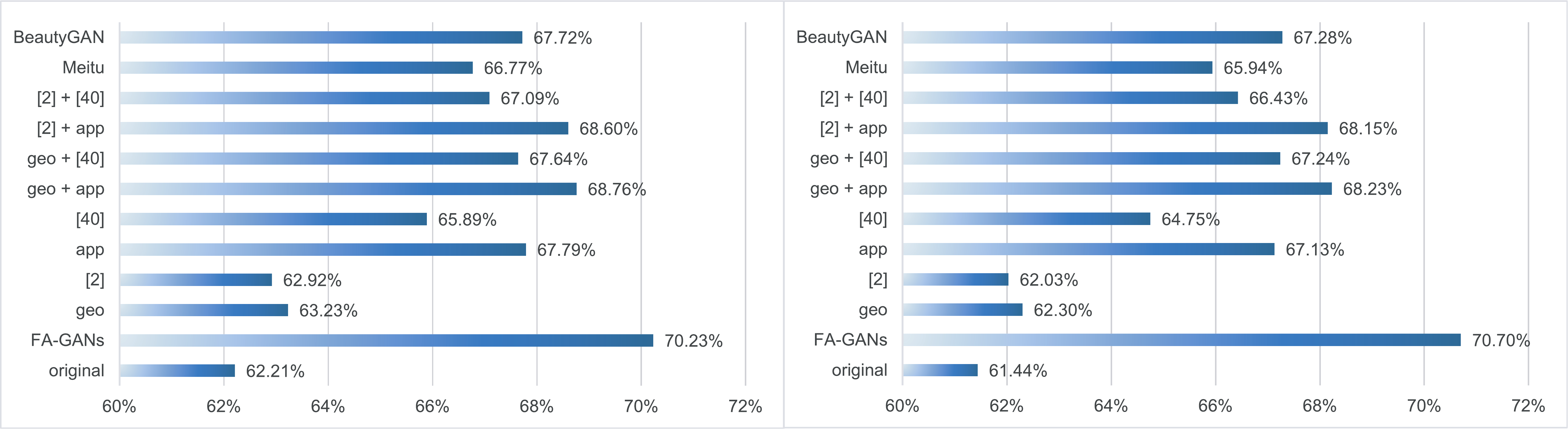}
\caption{Statistical results of beauty scores. We assess the facial attractiveness of original and the adjusted faces. The assessing results in female and male views are shown in the left and right, respectively.}	
\label{fig:statistical}	
\end{figure*}

%% file: files/tab_identity.tex
\begin{table*}
\centering
\caption{Identity preservation analysis on facial adjustment}
\label{tab:identity}
\begin{tabular}[t]{c c c c c c c c c c c c}
\hline
original & FA-GANs & geo & ~\cite{Leyvand:2008:DEF:1399504.1360637} & app & ~\cite{Li_2018_ECCV} & geo $+$ app & geo $+$ ~\cite{Li_2018_ECCV} & ~\cite{Leyvand:2008:DEF:1399504.1360637} $+$ app & ~\cite{Leyvand:2008:DEF:1399504.1360637} $+$ ~\cite{Li_2018_ECCV} & Meitu & BeautyGAN \\
\hline
$100.00\%$ & $93.91\%$ & $96.34\%$ & $94.67\%$ & $96.07\%$ & $84.27\%$ & $92.95\%$ & $82.55\%$ & $91.38\%$ & $81.57\%$ & $97.73\%$ & $93.46\%$ \\
\hline
\end{tabular}
\end{table*}

%% file: files/tab_time.tex
\begin{table}[t]
\centering
\caption{Time consuming analysis on geometry adjustment.}
\label{tab:geoAdjust}
\begin{tabular}[t]{c c c c c}
\hline
&$x \to l_x$ & $l_x \to l_y$ & $l_y \to y$ & total \\ \hline
Geo branch& $0.0779$ & $0.0039$ & $2.3039$ & $2.3857$ \\
~\cite{Leyvand:2008:DEF:1399504.1360637} & $0.0791$ & $0.0016$ & $2.1218$ & $2.2025$\\
\hline
\end{tabular}
\end{table}
\begin{table}
\centering
\caption{Time consuming analysis on appearance adjustment.}
\label{tab:appAdjust}
\begin{tabular}[t]{c c c c c}
\hline
&$x \to r$ & face parsing & $x \to y$ & total \\ \hline
App branch& $-$ & $-$ & $0.0599$ & $0.0599$ \\
~\cite{Li_2018_ECCV} & $0.0003$ & $0.0495$ & $6.2275$ & $6.2773$\\
\hline
\end{tabular}
\end{table}

\begin{table}[t]
\centering
\caption{Time consuming analysis on both geometry and appearance adjustment.}
\label{tab:twoStepAdjust}
\begin{tabular}[t]{c c c c c}
\hline
geo $+$ app & geo $+$~\cite{Li_2018_ECCV} & ~\cite{Leyvand:2008:DEF:1399504.1360637} $+$ app & ~\cite{Leyvand:2008:DEF:1399504.1360637} $+$ ~\cite{Li_2018_ECCV} & FA-GANs \\ \hline
$2.4456$ & $8.663$ & $2.2624$ & $8.4798$ & $\mathbf{0.0552}$ \\
\hline
\end{tabular}
\end{table}

%% file: files/Conclusions.tex
\section{Conclusions}
We have presented FA-GANs, a deep end-to-end framework for automating facial attractiveness enhancement in both the geometry and appearance aspects.
FA-GANs learn the implicit attractiveness rules via the pre-trained facial attractiveness ranking module and avoid training on the paired faces for which a large dataset is extremely difficult to obtain. 
In this way, FA-GANs enhance facial attractiveness in a data-driven manner.
FA-GANs contain two branches of geometry adjustment and appearance adjustment, and both of them can enhance the attractiveness independently.
FA-GANs generate compelling perceptual results and enhance facial attractiveness both effectively and efficiently. These are verified by our comprehensive experiments and thorough analysis, which also demonstrate that FA-GANs achieve superior performance over existing geometry and appearance enhancement methods.

Although we have shown the superiority of FA-GANs, there still exist several aspects that need to be improved in the future. 
FA-GANs are limited to the frontal faces, and this is because few faces of other poses are collected in our dataset. And these faces are much more difficult to collect than frontal faces.
A possible solution is that estimating the facial pose and adjusting the facial appearance and geometry in 3D face space. The attractiveness of the adjusted faces in other poses can also be evaluated.
Another possible solution is that enlarging the facial attractiveness dataset and training facial attractiveness enhancement networks involving lots of faces with variant poses.